\documentclass[10pt,twocolumn,letterpaper]{article}

\usepackage[pagenumbers]{cvpr}
\usepackage[accsupp]{axessibility}

\usepackage{graphicx}
\usepackage{amsmath}
\usepackage{amssymb}
\usepackage{booktabs}
\usepackage{tabularx}

\usepackage[pagebackref,breaklinks,colorlinks]{hyperref}
\usepackage{amsfonts}
\usepackage{nicefrac}
\usepackage{microtype}
\usepackage{xcolor}
\usepackage{lipsum}
\usepackage{wrapfig}
\graphicspath{{figures/}}

\newcommand{\xdiffmask}{\textsc{DiffMask}}
\newcommand{\diffmask}{\xdiffmask~}
\newcommand{\xvizdiffmask}{\textsc{Vision~DiffMask}}
\newcommand{\vizdiffmask}{\xvizdiffmask~}

\newcommand{\bb}{\mathbf{b}}
\newcommand{\bh}{\mathbf{h}}
\newcommand{\bx}{\mathbf{x}}
\newcommand{\by}{\mathbf{y}}
\newcommand{\bz}{\mathbf{z}}
\newcommand{\bu}{\mathbf{u}}

\newcommand{\bphi}{\boldsymbol\phi}

\newcommand{\norm}[1]{\left\lVert#1\right\rVert}

\newcommand{\citep}{\cite}
\newcommand{\citet}{\cite}

\usepackage[capitalize]{cleveref}
\crefname{section}{Sec.}{Secs.}
\Crefname{section}{Section}{Sections}
\Crefname{table}{Table}{Tables}
\crefname{table}{Tab.}{Tabs.}

\def\cvprPaperID{26}
\def\confName{CVPR}
\def\confYear{2023}

\begin{document}

\title{\xvizdiffmask: Faithful Interpretation of Vision\\Transformers with Differentiable Patch Masking}

\author{
Angelos Nalmpantis\thanks{Equal contribution} \thanks{Correspondence to: angelosnalm@gmail.com} , Apostolos Panagiotopoulos\footnotemark[1] , John Gkountouras\footnotemark[1] ,\\ Konstantinos Papakostas\footnotemark[1]\phantom{,} and Wilker Aziz \\[0.2cm]
University of Amsterdam
}

\maketitle

\begin{abstract}
    The lack of interpretability of the Vision Transformer may hinder its use in critical real-world applications despite its effectiveness. To overcome this issue, we propose a post-hoc interpretability method called \xvizdiffmask, which uses the activations of the model's hidden layers to predict the relevant parts of the input that contribute to its final predictions. Our approach uses a gating mechanism to identify the minimal subset of the original input that preserves the predicted distribution over classes. We demonstrate the faithfulness of our method, by introducing a faithfulness task, and comparing it to other state-of-the-art attribution methods on CIFAR-10 and ImageNet-1K, achieving compelling results. To aid reproducibility and further extension of our work, we open source our implementation \href{https://github.com/AngelosNal/Vision-DiffMask}{here}.
\end{abstract}

\vspace{-0.2cm}
\section{Introduction}
\label{sec:introduction}
The Vision Transformer (ViT)~\cite{dosovitskiy2020vit} has been a major breakthrough in recent years, with applications in tasks such as image classification, object detection, and image captioning. However, its success comes at the cost of interpretability, as deep neural networks are usually treated like ``black boxes'' that do not provide any insight into their decision-making process. This becomes a major drawback for many real-world applications that require safety and social acceptance.

As a result, there has been a surge of work in the area of \textit{post-hoc} interpretability, which aims to explain how models arrive at their decisions. These methods can be categorized into two main groups; \textit{model-specific} and \textit{model-agnostic} methods. Model-specific methods are tailored to a specific architecture and make use of a model's internal structure to generate explanations. They are usually faster and more efficient, but they are not applicable to other models, which makes them less flexible when a new architecture is proposed. On the other hand, model-agnostic methods can be applied to any model, but are often more computationally expensive and require more data to train.

\begin{figure}[t]
    \centering
    \includegraphics[width=\columnwidth, trim=0 19.25cm 0 0, clip]{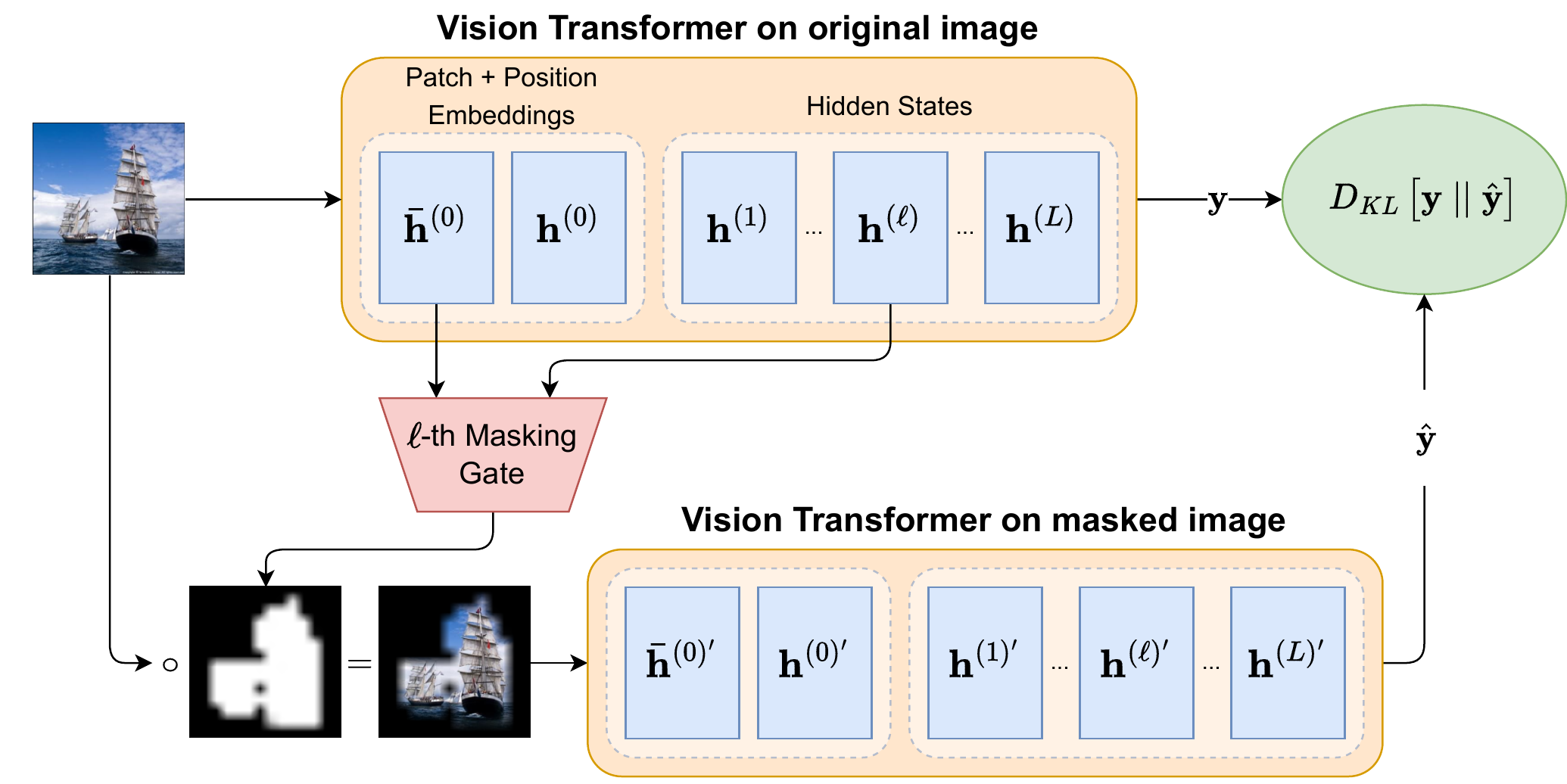}
    \caption{Overview of \xvizdiffmask's architecture.}
    \label{fig:model}
    \vspace{-0.2cm}
\end{figure}

Within the model-specific category, a natural consequence of the Transformer's~\cite{vaswani2017attention} self-attention mechanism is to use the attention weights to explain the predicted output. A popular method in this line of research is \textit{attention rollout}~\cite{abnar2020rollout}, which uses the attention weights to generate a saliency map over the input. However, a consensus has not been reached on whether attention can be considered as a faithful interpretation of the model's decision-making process, with advocates both in favor~\cite{wiegreffe2019attention-not-not} and against it~\cite{jain2019attention-not}.

On the other hand, model-agnostic methods are generally split into two sub-categories; \textit{gradient-based}, and \textit{attribution propagation} approaches. The former is characterized by the use of the gradients of a model's output with respect to a layer's input as an indicator of importance~\cite{simonyan14saliency, sundararajan2017integrated-gradients, selvaraju2017grad-cam}, while the latter relies on the Deep Taylor Decomposition method~\cite{montavon2017taylor} to recursively break down the model's output into the contributions of each layer~\cite{binder2016lrp,shrikumar2017deeplift,chefer2021interpretability}. Other approaches that do not fall into these two categories are methods based on input perturbations~\cite{fong2017perturbation, ribeiro2016should, Petsiuk2018rise, dabkowski2017real} or excitation back-propagation~\cite{zhang2018excitation}, with a common theme being the treatment of the model as a black box. However, this usually implies an increase in computational complexity, which makes them infeasible to use with large models like ViT.

A common shortcoming of these methods is that they do not guarantee that the model is fully ignoring low-scored features, or that the output distribution will be preserved in their absence. Hence, a question that occurs naturally is whether these explanations are \textit{faithful} to the model's decision-making process. Recent works have relied on \textit{erasure}~\cite{erasure} to tackle this issue, with De Cao et al.~\cite{decao2020diffmask} proposing \xdiffmask, an interpretation method for the language domain that predicts attribution spans over an input text based on a model's hidden representations. The training objective is designed to keep the minimal subset~\cite{bastings2019interpretable} of the input that produces a similar output distribution over the target labels.

In this paper, we extend \diffmask to the vision domain by introducing \xvizdiffmask, an interpretation network that predicts saliency maps for models that follow ViT's architecture. Our method fundamentally relies on a series of gating mechanisms on each of ViT's layers, which are optimized with the goal of preserving the model's output when masking the input. During training, all gates cast a binary vote on whether an image patch should be kept or not. Subsequently, the votes are aggregated across layers to produce the final mask. During inference, each gate predicts a probability instead of a binary vote, creating a continuous attribution map over the input. However, we empirically demonstrate that, due to our training objective, the attributions collapse to hard boundaries between relevant and irrelevant patches. This design choice allows us to confidently identify significant portions of the input that the model disregards during its predictions.

Attribution methods cannot be evaluated simply by using human annotations, for that would measure the plausibility of the explanations according to humans, and not a faithful attribution according to the model~\cite{jacovi2020faithfulness}. Hence, we first test the faithfulness of our model in a controlled scenario. Then, we evaluate our methodology using both qualitative and quantitative experiments against other state-of-the-art methods on CIFAR-10~\cite{krizhevsky2009cifar} and ImageNet-1K~\cite{den2009imagenet}. We show that our method produces faithful and plausible outputs and can be used to provide insights into the model's inner workings.

In summary, our main contributions are:
\begin{enumerate}
    \item We introduce \xvizdiffmask, a novel method for post-hoc interpretability in the vision domain.
    \item We formulate a faithfulness task and show that our proposed method is indeed faithful to the model's decision-making process.
    \item We evaluate our method on CIFAR-10 and ImageNet-1K, both qualitatively and quantitatively, and achieve strong results against common interpretation methods.

\end{enumerate}

\section{Methodology}
\label{sec:methodology}

\vizdiffmask is applied on a pre-trained Vision Transformer and produces a saliency map for each input image $\bx$. During the forward pass, we transform each of the model's hidden states into a patch-level mask. These masks are then aggregated over the hidden layers and result in the final mask $\bz$ that is applied to the input. An overview of our approach is shown in Figure~\ref{fig:model}.

\subsection{Gating Mechanism}
\label{sec:gates}
Our method uses a total of $L+2$ gates to transform the hidden states to masks; the first two process the patch embeddings, $\bar{\bh}^{(0)}$ and $\bh^{(0)}$, without and with the positional embeddings added respectively, while the rest $L$ process ViT's hidden states $\left(\bh^{(\ell)}, \ell \in \{1, 2, \ldots, L\}\right)$. For each state $\bh$, the input to the corresponding gate is the concatenation $[\bar{\bh}^{(0)}; \bh]$. We use MLPs with $\tanh$ activations as gating mechanisms, followed by a linear stretch. This stretch gives us explicit control over the range of probabilities in the saliency map, which allows us to choose a suitable percentage of image patches to be masked during initialization. This results in a total of $L+2$ patch-level activations for each image:
\begin{equation}
    \bu^{(\ell)} = \alpha \cdot \text{MLP}([\bar{\bh}^{(0)}; \bh^{(\ell)}]) + \beta \cdot \mathbf{1} ~,
    \label{eq:gates}
\end{equation}
where $\alpha, \beta$ are the parameters of the linear stretch.

\subsection{Mask Generation}
The activations $\bu^{(\ell)}$ take values in the $(-\infty, \infty)$ range which is unsuitable for saliency maps. One way to convert them to the $[0, 1]$ range would be to apply a softmax, but this would assign no probability mass to truly mask out an image patch. To overcome this, we use $\bu^{(\ell)}$ as parameters for the modified Hard Concrete (HC) distribution~\citep{louizos2018hardconcrete,decao2020diffmask}, which gives support to $[0,1)$. Hence, during training, our masks are sampled as follows:
\begin{align}
    \bz^{(\ell)} &\sim \text{HardConcrete}(\bz^{(\ell)}; \bu^{(\ell)}, l, r)
\end{align}
where $l\leq 0$ and $r\geq 1$ control the probability density of the HC distribution at $0$ and $1$. During inference, instead of sampling $\bz^{(\ell)}$, we take its expected value. The reason for this is that during training we need \vizdiffmask to understand the importance of each patch while during inference we want a smoother attribution map over the input. We calculate the mask of $\bx$ as the product $\bz = \prod \bz^{(\ell)}$. As this results in a patch-level saliency map, we use bi-linear interpolation to arrive at a higher granularity pixel-level map.

If we were to naively multiply $\bx$ with $\bz$, we would be replacing masked-out patches with zeros, which corresponds to a black patch. Although this makes sense from a human's perspective, ViT has not learned to ignore these patches but rather to treat them as another source of information. To address this issue, we use a special vector $\bb$ that is learned along with the parameters of the gating mechanisms and is shared across all instances in the dataset. Ultimately, we define the \textit{element-wise} masking operation as:
\begin{equation}
    \label{eq:masking}
    \hat{\bx}_i = z_i \cdot \bx_i + (1 - z_i) \cdot \bb
\end{equation}

\subsection{Training Objective}
We cannot rely on human annotations to train \xvizdiffmask, so we resort to a self-supervised objective. We train our interpretation module to produce a mask, such that the masked input gives an output sufficiently close to the original one, while simultaneously ignoring as many patches as possible. In the case of image classification, the first objective is achieved by introducing the KL divergence $D_{\operatorname{KL}}[\by \mid\mid \hat{\by}]$ between the unmasked $\by$ and the masked predictions $\hat{\by}$ in the loss. The second one is achieved by including the $L_0$ norm\footnote{The $L_0$ norm of a vector is defined as the number of non-zero elements, i.e. $\norm{z}_0 = \#\{i \mid z_i \neq 0\}$. Although it is not a proper norm from a mathematical perspective, it is often referred to as such in the literature.} of the predicted mask in the loss, denoted as $\mathcal{L}_0 (\bphi, \bb \mid \bx)$, where $\bphi$ are the parameters of the gates and $\bb$ is the learnable baseline vector of Eq.~\ref{eq:masking}.

The described training procedure can be expressed as a constrained optimization problem, where we minimize $\mathcal{L}_0$ -- and thus aim to mask out as many patches as possible -- while keeping the divergence of the two distributions $D_{\operatorname{KL}}$ within an acceptable margin $m$:
\begin{equation}
    \underset{\bphi, \bb}{\min} \sum_{\bx \in \mathcal{D}} \mathcal{L}_0 (\bphi, \bb \mid \bx) \quad \mathrm{s.t.}~ D_{\operatorname{KL}}[\by \mid\mid \hat{\by}] \leq m,~
\end{equation}
To deal with the intractability of non-linear constraint optimizations, we can equivalently express this as a Lagrangian relaxation problem~\cite{boyd2004optim}:
\begin{equation}
    \underset{\lambda}{\max}~ \underset{\bphi, \bb}{\min} \sum_{\bx \in \mathcal{D}} \mathcal{L}_0 (\bphi, \bb \mid \bx) + \lambda \left( D_{\operatorname{KL}}[\by \mid\mid \hat{\by}] - m \right)
\end{equation}
where $\lambda \geq 0$ is the Lagrangian multiplier.

\section{Experiments}
\label{sec:experiments}

\subsection{Implementation \& Training Details}
\label{sec:training_details}
We employ a \vizdiffmask architecture with $14$ gates for all of our experiments, with each gate being a two-layer MLP using a $\tanh$ activation. For CIFAR, we use a linear stretch with $\alpha = 15$ and $\beta = 8$ (in Eq.~\ref{eq:gates}).
By doing so, the expected percentage of masked patches in a randomly initialized model is $\sim30\%$. Furthermore, we initialize $\lambda$ to $20$ and $m$ to $0.1$, and we use $3$ different learning rates to optimize our model: $2\cdot10^{-5}$ for the gates' parameters, $10^{-3}$ for the baseline vector, and $0.3$ for the Lagrangian. Additionally, we used a batch size of $16$ and trained for $25$ epochs. Finally, we chose the LookAhead Adam optimizer~\citep{zhang2019lookahead}. For ImageNet, we initialize our model with the converged model on CIFAR. In this case, we set the linear stretch to $\alpha = 20$, and adjust the learning rates for the gates' parameters to $10^{-5}$, for the baseline vector to $5\cdot 10^{-3}$, and for the Lagrangian to $0.15$. We provide additional information with our training observations and computational requirements in Appendices \ref{app:training} \& \ref{app:comp_requirements}.

\subsection{Faithfulness Task: Counting Patches}
In order to verify that our method is indeed faithful, we evaluate it on a problem where we know the ground truth by design, such as by counting the appearances of an object of interest. To this end, we propose the following setup: on a $3\times3$ grid, we color $n$ random patches with red, and the rest with a random combination of $6$ other colors. We then train a ViT with the goal of counting the number of red patches on a given grid, and, as expected, achieve a 100\% accuracy.\footnote{More details about the faithfulness task can be found in Appendix~\ref{app:faith_task}.}

\begin{figure}[tb]
    \vspace{-0.7cm}
    \centering
    \includegraphics[width=\columnwidth]{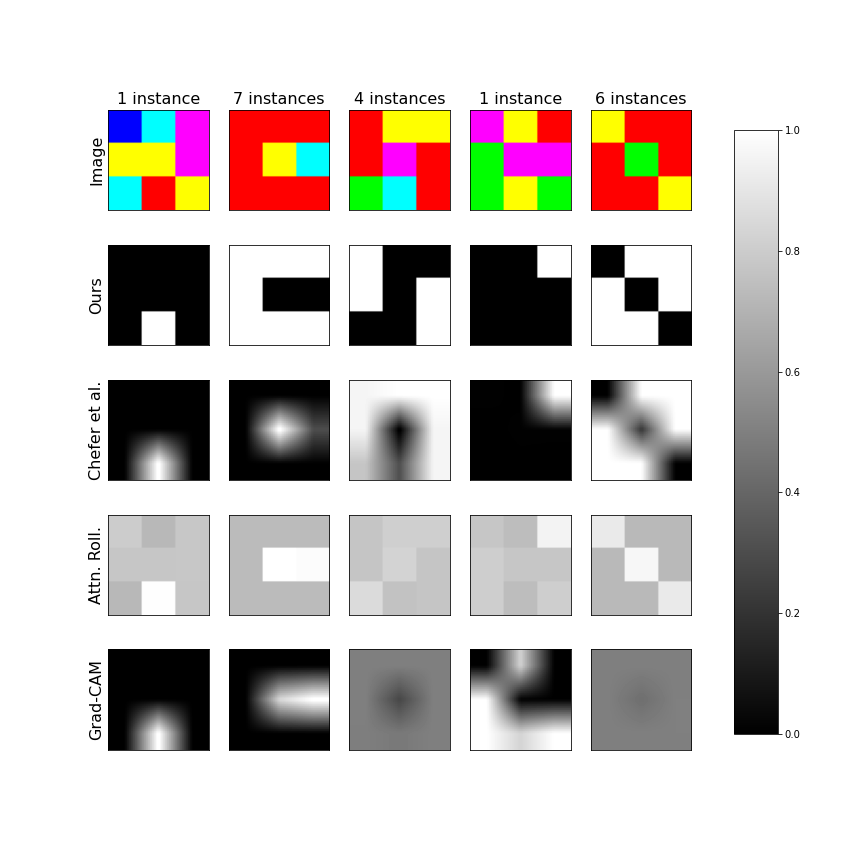}
    \vspace{-1.5cm}
    \caption{Saliency maps of \vizdiffmask and rival methods on the faithfulness task. As the model counts the red patches on the grid, it only needs to inspect either the red subset or its complement.}
    \label{fig:toy_task}
    \vspace{-0.5cm}
\end{figure}

We expect the decision-making process in ViT to involve either only the red-colored patches in the grid or all the others; any more is not needed to arrive at a correct prediction and any less is not sufficient to correctly solve the task. We compare our approach to well-established methods, such as the one proposed by Chefer et al.~\cite{chefer2021interpretability}, Attention Rollout~\cite{abnar2020rollout}, and Grad-CAM~\cite{selvaraju2017grad-cam}. As shown in Figure~\ref{fig:toy_task}, \vizdiffmask is the only method that produces \textit{faithful} and consistent interpretations for all inputs, while rival methods occasionally fail to do so.

\subsection{Image Classification}
\begin{figure}[t]
    \centering
    \includegraphics[width=\columnwidth]{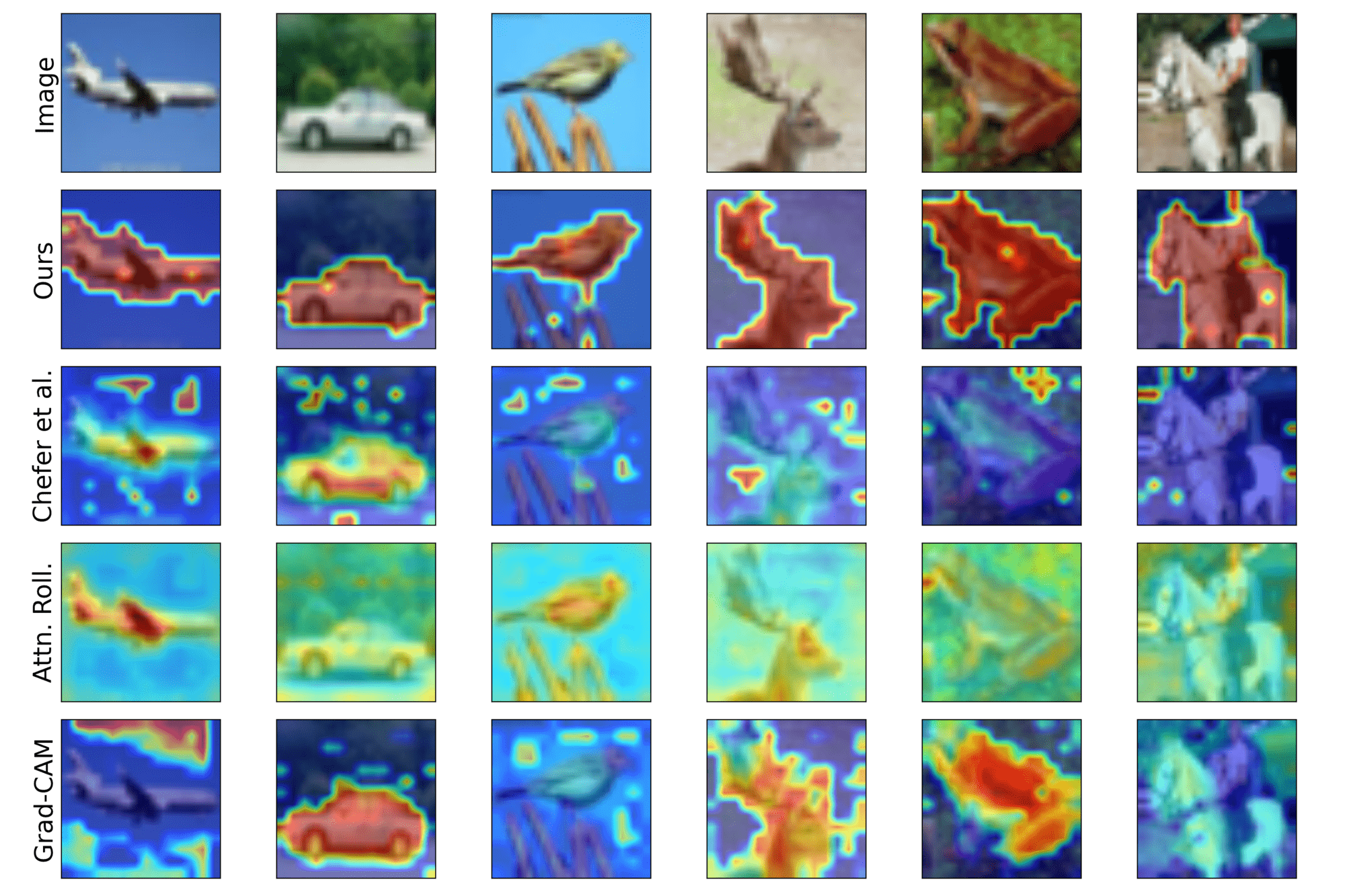}
    \caption{Attribution maps of \vizdiffmask and rival methods on sample images from CIFAR-10.}
    \label{fig:all_methods_cifar}
    \vspace{-0.4cm}
\end{figure}

\begin{figure}[t]
    \centering
    \includegraphics[width=\columnwidth]{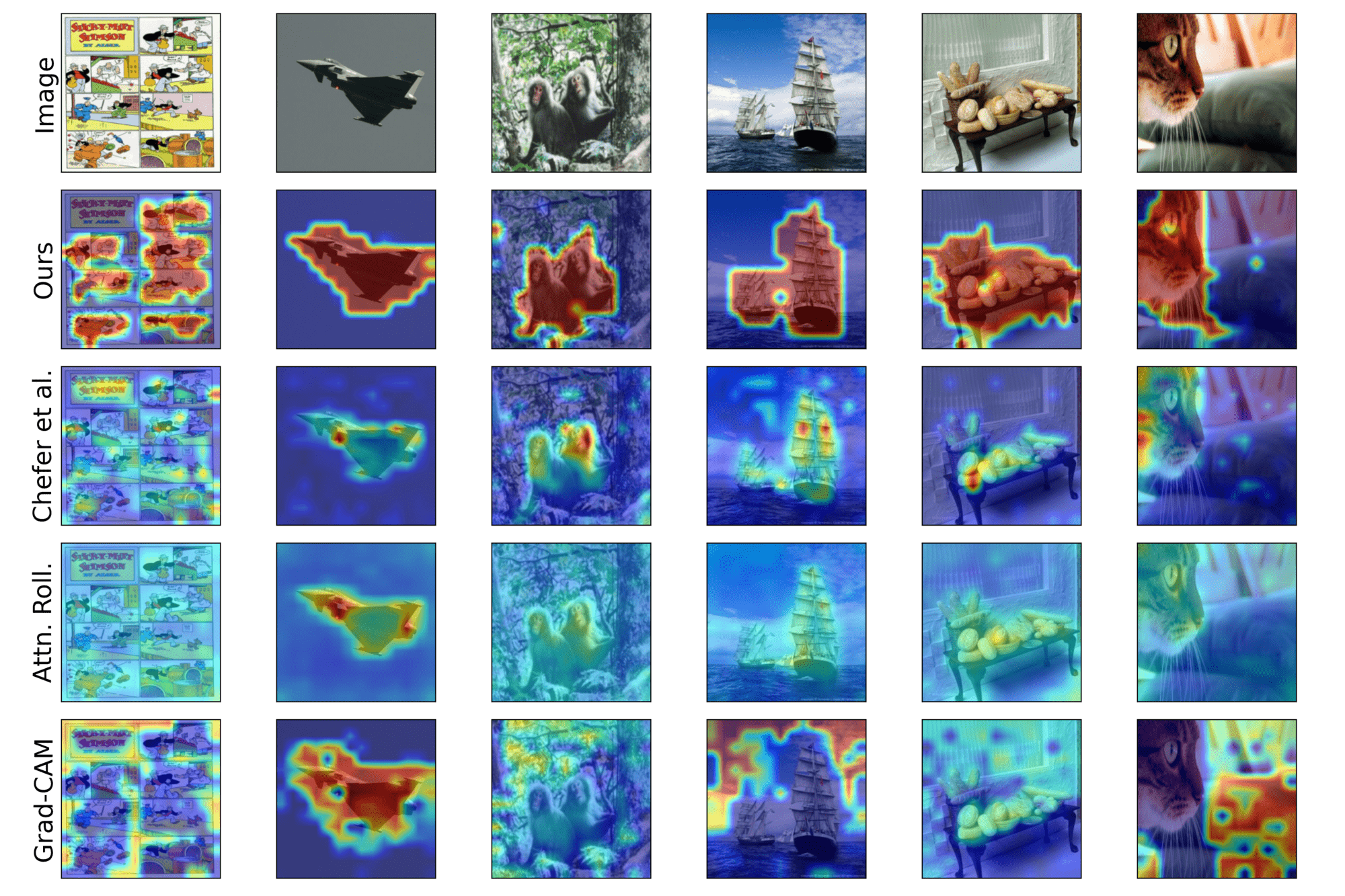}
    \caption{Attribution maps of \vizdiffmask and rival methods on sample images from ImageNet-1K.}
    \label{fig:all_methods_imagenet}
\end{figure}

 We evaluate our method on the standard setting of image classification, using the CIFAR-10 and ImageNet-1K datasets. We use ViT models with $98.75\%$ and $85.49\%$ accuracy respectively, and plot the saliency maps produced by \vizdiffmask and other methods in Figures \ref{fig:all_methods_cifar} \& \ref{fig:all_methods_imagenet}.

In CIFAR-10, objects are typically centered and occupy a significant portion of the image, suggesting that \vizdiffmask should primarily focus on the central object within an image. This hypothesis is supported by the saliency maps presented in Figure~\ref{fig:all_methods_cifar}, which consistently attribute the primary object. The ImageNet-1K dataset presents greater complexity due to its increased number of classes and frequent inclusion of multiple objects within a single image. As our method is trained to preserve the entire class distribution of the input rather than solely its top class, we would anticipate attributions across all prominent objects within an image. This expectation is confirmed in Figure~\ref{fig:all_methods_imagenet}.

Our findings demonstrate that \vizdiffmask effectively attributes the primary object(s) within the images, aligning with the current literature on result plausibility. In contrast, competing methods attribute regions that seem unlikely to contribute to the model's decision. A key distinction of \vizdiffmask is its ability to delineate clear boundaries between what it identifies as the \textit{triggering subset} and the remainder of the image. Despite adapting our training objective from the language domain, spatial continuity is largely maintained within highly attributed patches. However, it should be noted that our method occasionally omits low-frequency patches that are part of the primary object while retaining secondary objects within the image (e.g., a person riding a horse in the rightmost image of Figure~\ref{fig:all_methods_cifar}).

Although qualitative analysis of results can provide valuable insights into interpretability methods, there is no straightforward approach to evaluate them~\cite{decao2020diffmask}. Hence, to complement the above analysis, we also calculated the positive and negative perturbation curves, proposed by Chefer et al.~\cite{chefer2021interpretability}.
These curves show how the KL divergence and top-1 accuracy vary when removing pixels or patches with increasing and decreasing order of importance according to the model's attributions. We report the AUC of those curves for CIFAR-10 in Table~\ref{tab:quatitative} and provide the full plots in Appendix~\ref{app:curves}. \vizdiffmask surpasses all other methods when removing image patches with \textit{decreasing} order of importance (neg. perturbation), while remaining competitive when removing patches with \textit{increasing} order of importance (pos. perturbation). This is expected as our method's attributions collapse to hard boundaries, giving no relative attributions to patches. For example, in the airplane image in Figure~\ref{fig:all_methods_cifar}, our method considers the whole airplane to trigger the model's prediction, while Attention Rollout mostly focuses on the fins. This causes the latter to remove them first, which likely deteriorates ViT's performance more than an ostensibly random plane pixel removed by our method.

\begin{table}[t]
    \resizebox{\columnwidth}{!}{%
    \begin{tabular}{@{}lcccc@{}}
    \toprule
    Metric for AUC & Ours & Grad-CAM & Attn. Roll. & Chefer et al. \\ \midrule
    Pos. $D_{\operatorname{KL}}$ $(\uparrow)$ & 39.3 & 43.6 & 23.5 & \textbf{50.5} \\
    Neg. $D_{\operatorname{KL}}$ $(\downarrow)$ & \textbf{17.4} & 37.8 & 26.1 & 31.2 \\
    Pos. accuracy $(\downarrow)$ & 12.9 & 12.6 & 18.8 & \textbf{9.8} \\
    Neg. accuracy $(\uparrow)$ & \textbf{21.8} & 14.7 & 17.8 & 17.1 \\ \bottomrule
    \end{tabular}%
    }
    \caption{Quantitative comparison of \vizdiffmask and rival methods on CIFAR-10. $\uparrow$ ($\downarrow$) indicates that higher (lower) is better. \textbf{Bold} numbers correspond to the best result in each row.}
    \label{tab:quatitative}
    \vspace{-0.5cm}
\end{table}

\section{Conclusion}
\label{sec:conclusion}

In this work, we introduced \xvizdiffmask, a post-hoc interpretation module for the Vision Transformer. Our method predicts a mask that preserves only the minimal subset of patches and when applied results in the same output distribution as the original image.
We proposed a faithfulness task and demonstrated that \vizdiffmask provides faithful interpretations while other approaches fail to do so consistently. We also evaluated our method on two image classification datasets, CIFAR-10 and ImageNet-1K, and achieved plausible interpretations, with compelling quantitative metrics.
With our study, we exhibit the lack of faithfulness tasks in the field of Explainable AI and aim to pave the way for more sophisticated interpretation methods that are thoroughly evaluated for their faithfulness.

\section*{Acknowledgements}

\begin{wrapfigure}[3]{l}{0.10\linewidth}
\vspace{-13pt}
\includegraphics[width=0.08\textwidth]{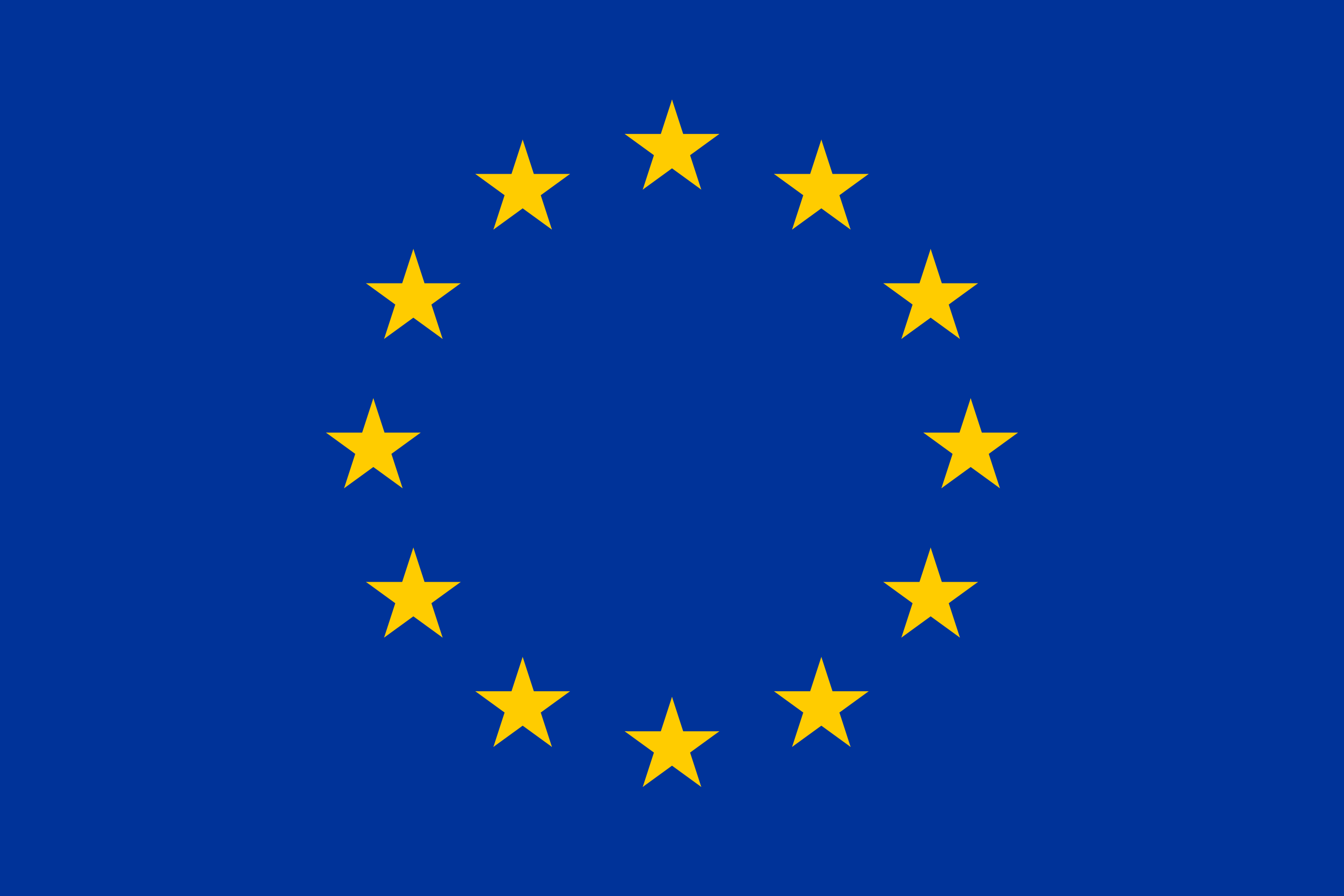}
\end{wrapfigure}
\noindent This work received funding from the EU's Horizon Europe RIA (UTTER, contract 101070631).

{\small \noindent
\bibliographystyle{ieee_fullname}
\bibliography{main}

\begin{thebibliography}{10}\itemsep=-1pt

\bibitem{abnar2020rollout}
Samira Abnar and Willem Zuidema.
\newblock Quantifying attention flow in transformers.
\newblock In {\em Proceedings of the 58th Annual Meeting of the Association for
  Computational Linguistics}, pages 4190--4197, Online, July 2020. Association
  for Computational Linguistics.

\bibitem{bastings2019interpretable}
Jasmijn Bastings, Wilker Aziz, and Ivan Titov.
\newblock Interpretable neural predictions with differentiable binary
  variables.
\newblock In {\em Proceedings of the 57th Annual Meeting of the Association for
  Computational Linguistics}, pages 2963--2977, Florence, Italy, July 2019.
  Association for Computational Linguistics.

\bibitem{binder2016lrp}
Alexander Binder, Gr{\'e}goire Montavon, Sebastian Lapuschkin, Klaus-Robert
  M{\"u}ller, and Wojciech Samek.
\newblock Layer-wise relevance propagation for neural networks with local
  renormalization layers.
\newblock In Alessandro~E.P. Villa, Paolo Masulli, and Antonio~Javier
  Pons~Rivero, editors, {\em Artificial Neural Networks and Machine Learning --
  ICANN 2016}, pages 63--71, Cham, 2016. Springer International Publishing.

\bibitem{boyd2004optim}
Stephen Boyd, Stephen~P Boyd, and Lieven Vandenberghe.
\newblock {\em Convex optimization}.
\newblock Cambridge university press, 2004.

\bibitem{chefer2021interpretability}
Hila Chefer, Shir Gur, and Lior Wolf.
\newblock Transformer interpretability beyond attention visualization.
\newblock In {\em Proceedings of the IEEE/CVF Conference on Computer Vision and
  Pattern Recognition (CVPR)}, pages 782--791, June 2021.

\bibitem{dabkowski2017real}
Piotr Dabkowski and Yarin Gal.
\newblock Real time image saliency for black box classifiers.
\newblock In I. Guyon, U.~Von Luxburg, S. Bengio, H. Wallach, R. Fergus, S.
  Vishwanathan, and R. Garnett, editors, {\em Advances in Neural Information
  Processing Systems}, volume~30. Curran Associates, Inc., 2017.

\bibitem{decao2020diffmask}
Nicola De~Cao, Michael~Sejr Schlichtkrull, Wilker Aziz, and Ivan Titov.
\newblock How do decisions emerge across layers in neural models?
  interpretation with differentiable masking.
\newblock In {\em Proceedings of the 2020 Conference on Empirical Methods in
  Natural Language Processing (EMNLP)}, pages 3243--3255, Online, Nov. 2020.
  Association for Computational Linguistics.

\bibitem{den2009imagenet}
Jia Deng, Wei Dong, Richard Socher, Li-Jia Li, Kai Li, and Li Fei-Fei.
\newblock Imagenet: A large-scale hierarchical image database.
\newblock In {\em 2009 IEEE Conference on Computer Vision and Pattern
  Recognition}, pages 248--255, 2009.

\bibitem{dosovitskiy2020vit}
Alexey Dosovitskiy, Lucas Beyer, Alexander Kolesnikov, Dirk Weissenborn,
  Xiaohua Zhai, Thomas Unterthiner, Mostafa Dehghani, Matthias Minderer, Georg
  Heigold, Sylvain Gelly, Jakob Uszkoreit, and Neil Houlsby.
\newblock An image is worth 16x16 words: Transformers for image recognition at
  scale.
\newblock In {\em International Conference on Learning Representations}, 2021.

\bibitem{fong2017perturbation}
Ruth~C. Fong and Andrea Vedaldi.
\newblock Interpretable explanations of black boxes by meaningful perturbation.
\newblock In {\em Proceedings of the IEEE International Conference on Computer
  Vision (ICCV)}, Oct 2017.

\bibitem{jacovi2020faithfulness}
Alon Jacovi and Yoav Goldberg.
\newblock Towards faithfully interpretable {NLP} systems: How should we define
  and evaluate faithfulness?
\newblock In {\em Proceedings of the 58th Annual Meeting of the Association for
  Computational Linguistics}, pages 4198--4205, Online, July 2020. Association
  for Computational Linguistics.

\bibitem{jain2019attention-not}
Sarthak Jain and Byron~C. Wallace.
\newblock {A}ttention is not {E}xplanation.
\newblock In {\em Proceedings of the 2019 Conference of the North {A}merican
  Chapter of the Association for Computational Linguistics: Human Language
  Technologies, Volume 1 (Long and Short Papers)}, pages 3543--3556,
  Minneapolis, Minnesota, June 2019. Association for Computational Linguistics.

\bibitem{krizhevsky2009cifar}
Alex Krizhevsky.
\newblock Learning multiple layers of features from tiny images.
\newblock Technical report, Canadian Institute for Advanced Research (CIFAR),
  2009.

\bibitem{erasure}
Jiwei Li, Will Monroe, and Dan Jurafsky.
\newblock Understanding neural networks through representation erasure.
\newblock {\em arXiv preprint arXiv:1612.08220}, 2016.

\bibitem{louizos2018hardconcrete}
Christos Louizos, Max Welling, and Diederik~P. Kingma.
\newblock Learning sparse neural networks through $l_0$ regularization.
\newblock In {\em International Conference on Learning Representations}, 2018.

\bibitem{montavon2017taylor}
Grégoire Montavon, Sebastian Lapuschkin, Alexander Binder, Wojciech Samek, and
  Klaus-Robert Müller.
\newblock Explaining nonlinear classification decisions with deep taylor
  decomposition.
\newblock {\em Pattern Recognition}, 65:211--222, 2017.

\bibitem{Petsiuk2018rise}
Vitali Petsiuk, Abir Das, and Kate Saenko.
\newblock Rise: Randomized input sampling for explanation of black-box models.
\newblock In {\em British Machine Vision Conference (BMVC)}, 2018.

\bibitem{ribeiro2016should}
Marco~Tulio Ribeiro, Sameer Singh, and Carlos Guestrin.
\newblock "why should i trust you?": Explaining the predictions of any
  classifier.
\newblock In {\em Proceedings of the 22nd ACM SIGKDD International Conference
  on Knowledge Discovery and Data Mining}, KDD '16, page 1135–1144, New York,
  NY, USA, 2016. Association for Computing Machinery.

\bibitem{selvaraju2017grad-cam}
Ramprasaath~R. Selvaraju, Michael Cogswell, Abhishek Das, Ramakrishna Vedantam,
  Devi Parikh, and Dhruv Batra.
\newblock Grad-cam: Visual explanations from deep networks via gradient-based
  localization.
\newblock In {\em Proceedings of the IEEE International Conference on Computer
  Vision (ICCV)}, Oct 2017.

\bibitem{shrikumar2017deeplift}
Avanti Shrikumar, Peyton Greenside, and Anshul Kundaje.
\newblock Learning important features through propagating activation
  differences.
\newblock In Doina Precup and Yee~Whye Teh, editors, {\em Proceedings of the
  34th International Conference on Machine Learning}, volume~70 of {\em
  Proceedings of Machine Learning Research}, pages 3145--3153. PMLR, 06--11 Aug
  2017.

\bibitem{simonyan14saliency}
Karen Simonyan, Andrea Vedaldi, and Andrew Zisserman.
\newblock Deep inside convolutional networks: Visualising image classification
  models and saliency maps.
\newblock In {\em In Workshop at International Conference on Learning
  Representations}, 2014.

\bibitem{sundararajan2017integrated-gradients}
Mukund Sundararajan, Ankur Taly, and Qiqi Yan.
\newblock Axiomatic attribution for deep networks.
\newblock In Doina Precup and Yee~Whye Teh, editors, {\em Proceedings of the
  34th International Conference on Machine Learning}, volume~70 of {\em
  Proceedings of Machine Learning Research}, pages 3319--3328. PMLR, 06--11 Aug
  2017.

\bibitem{vaswani2017attention}
Ashish Vaswani, Noam Shazeer, Niki Parmar, Jakob Uszkoreit, Llion Jones,
  Aidan~N Gomez, \L~ukasz Kaiser, and Illia Polosukhin.
\newblock Attention is all you need.
\newblock In I. Guyon, U.~Von Luxburg, S. Bengio, H. Wallach, R. Fergus, S.
  Vishwanathan, and R. Garnett, editors, {\em Advances in Neural Information
  Processing Systems}, volume~30. Curran Associates, Inc., 2017.

\bibitem{wiegreffe2019attention-not-not}
Sarah Wiegreffe and Yuval Pinter.
\newblock Attention is not not explanation.
\newblock In {\em Proceedings of the 2019 Conference on Empirical Methods in
  Natural Language Processing and the 9th International Joint Conference on
  Natural Language Processing (EMNLP-IJCNLP)}, pages 11--20, Hong Kong, China,
  Nov. 2019. Association for Computational Linguistics.

\bibitem{zhang2018excitation}
Jianming Zhang, Sarah~Adel Bargal, Zhe Lin, Jonathan Brandt, Xiaohui Shen, and
  Stan Sclaroff.
\newblock Top-down neural attention by excitation backprop.
\newblock In {\em Computer Vision -- ECCV 2016}, pages 543--559, Cham, 2016.
  Springer International Publishing.

\bibitem{zhang2019lookahead}
Michael Zhang, James Lucas, Jimmy Ba, and Geoffrey~E Hinton.
\newblock Lookahead optimizer: k steps forward, 1 step back.
\newblock In H. Wallach, H. Larochelle, A. Beygelzimer, F. d\textquotesingle
  Alch\'{e}-Buc, E. Fox, and R. Garnett, editors, {\em Advances in Neural
  Information Processing Systems}, volume~32. Curran Associates, Inc., 2019.

\end{thebibliography}
}

\clearpage
\appendix
\section{Perturbation Curves}
\label{app:curves}
\begin{figure}[ht]
     \centering
     \begin{subfigure}[b]{0.49\columnwidth}
         \centering
         \includegraphics[width=\columnwidth]{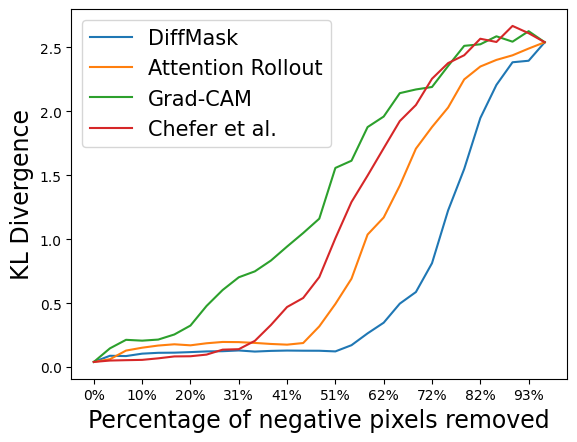}
     \end{subfigure}
     \hfill
     \begin{subfigure}[b]{0.49\columnwidth}
         \centering
         \includegraphics[width=\columnwidth]{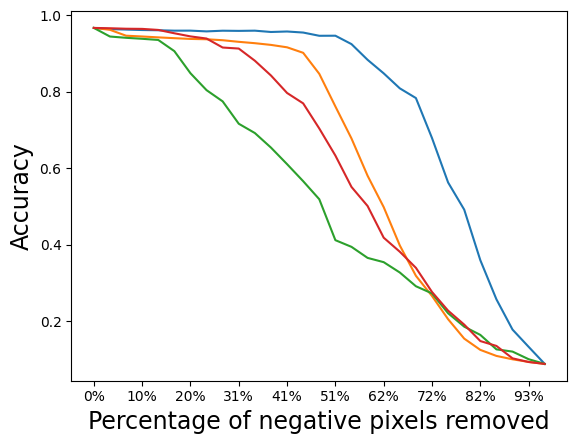}
     \end{subfigure}
        \caption{\textbf{Negative} perturbation results. Ideally, when removing pixels with a low attribution score, the model's output should remain the same, preserving a small KL divergence / high accuracy.}
        \label{fig:neg_perturb}
\end{figure}

\begin{figure}[ht]
     \centering
     \begin{subfigure}[b]{0.49\columnwidth}
         \centering
         \includegraphics[width=\columnwidth]{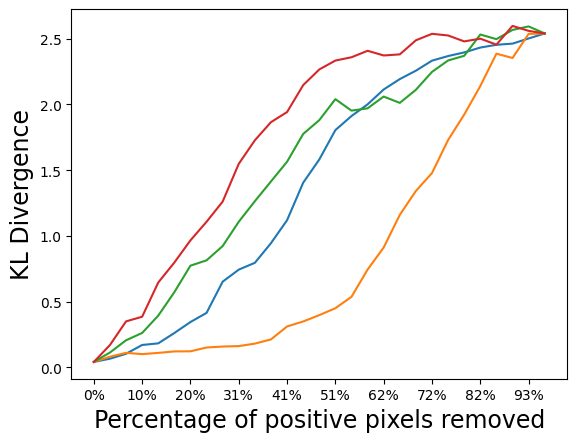}
     \end{subfigure}
     \hfill
     \begin{subfigure}[b]{0.49\columnwidth}
         \centering
         \includegraphics[width=\columnwidth]{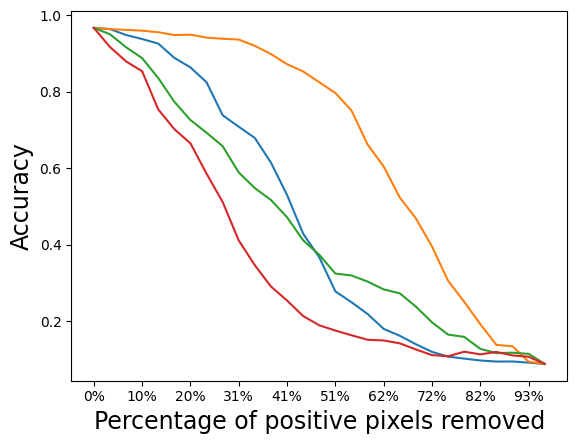}
     \end{subfigure}
        \caption{\textbf{Positive} perturbation results. Ideally, when removing pixels with a high attribution score, the model's output should radically change, leading to high KL divergence / low accuracy.}
        \label{fig:pos_perturb}
\end{figure}

We report the negative and positive perturbation curves in Figures~\ref{fig:neg_perturb} and~\ref{fig:pos_perturb} respectively. In addition to computing the drop in top-1 accuracy as in \cite{chefer2021interpretability}, we also compute the increase in KL divergence ($D_{\operatorname{KL}}$). Figure~\ref{fig:neg_perturb} showcases that our model indeed gives low attribution to the most irrelevant patches of the image, since removing them does not affect its performance, measured either by the top-1 accuracy or the divergence between the original and the new class distribution. However, since our method creates hard boundaries for what it believes to be the triggering subset of the image and the rest of the patches, removing the most important image patches does not impact the model's performance as much as in methods with progressive attribution, as shown in Figure~\ref{fig:pos_perturb}.

\section{Training Observations}
\label{app:training}
To aid the reproducibility of our work as well as encourage further research, we are apposing a few critical observations for the training procedure of \xvizdiffmask.

\subsection{General comments}
\label{app:initial_masking}
Even with the best hyperparameters, different random initializations can result in either convergent or divergent models. We attribute this to the fact that the initial few epochs are critical to the convergence of the model. If the initialization of the \vizdiffmask model creates an initial mask that results in a high KL divergence, the model has to precipitously explore the masking space. Otherwise, the optimizer for the Lagrange multiplier will overcompensate in either direction to lower the loss, which diverges the model. A possible solution to this would be to search for a method to controllably adjust the position of the initial mask. This is currently an indirect process as we can only control \xvizdiffmask's initialization.

\subsection{Effects of the linear stretch on convergence}
We chose the hyper-parameters of the linear stretch ($\alpha,\beta$) after an extensive search. Our configuration for CIFAR-10 results in randomly initialized gates masking out $\sim30\%$ of the input image's pixels. If that percentage was higher, which is the case when we are not stretching the output of the MLP for example, the model would not be able to learn image concepts and the baseline vector $\mathbf{b}$. This would render the KL divergence incapable of being reduced, leading to $\mathcal{L}_0$ being the only factor to optimize. We observed that this led to local minima where the entire image was being masked.

Given a \textit{small} $\beta$, the model will excessively mask the input, making it impossible to lower the loss unless the optimizer slowly decreases the Lagrangian multiplier to $0$. In turn, when $\lambda$ decreases to $0$, the optimization objective reduces to
\begin{equation}
    \underset{\bphi, \bb}{\min} \sum_{\bx \in \mathcal{D}} \mathcal{L}_0 (\bphi, \bb \mid \bx)
\end{equation}
which has an obvious optimum at masking the whole image, leading to divergence.

Given a \textit{large} $\beta$, the model will not mask anything at all. In turn, to keep the KL divergence constant near $0$, the optimizer will increase the $\alpha$, completely overpowering the masking objective, which again results in a divergent model.

\section{Faithfulness Task}
\label{app:faith_task}
For the faithfulness task of counting colored patches in a grid, we train a Vision Transformer using the HuggingFace implementation, for a \textit{maximum} of $20$ epochs with a batch size of $16$. We use a \texttt{ViTFeatureExtractor} that normalizes each of the image's channels to be centered around $0.5$ and have a standard deviation of $\pm0.5$. The optimizer used is \texttt{AdamW} with a learning rate of $5 \cdot 10^{-5}$ and a weight decay of $10^{-2}$. We use a scheduler that linearly decreases the learning rate after each epoch so that it reaches 0 at the end of $20$ epochs. However, we also resort to early stopping in case the validation set accuracy does not increase after $5$ epochs. For our synthetic dataset, this meant that training stopped after $8$ epochs. As one would expect for such a simple task, the Vision Transformer achieves 100\% accuracy on a hold-out evaluation set.

\section{Computational Requirements}
\label{app:comp_requirements}
All of our models were trained on 6 cores of an Intel Xeon Silver 4110 CPU @ $2.10$GHz and an NVIDIA GeForce GTX 1080 Ti GPU.
Training a single \vizdiffmask model for the image classification task on CIFAR-10, using a ViT-base model with $12$-layers and $16\times 16$ patches, required $10$ hours for the $25$ epochs it took to converge. Regarding the faithfulness task, pre-training a ViT-base model required $1$ minute for a total of $8$ epochs, while training the corresponding \vizdiffmask model required $9$ minutes for a total of $100$ epochs. Finally, the ImageNet model was fine-tuned for one epoch, with a duration of $18$ hours.
\end{document}